\documentclass[conference]{IEEEtran}
\usepackage{amsmath,amssymb,amsfonts}
\usepackage{algorithmic}
\usepackage{graphicx}
\usepackage{textcomp}
\usepackage{xcolor}
\usepackage[ruled,vlined]{algorithm2e}
\usepackage{booktabs}
\usepackage{multirow}
\usepackage{cite}
\usepackage[utf8]{inputenc}
\usepackage[T1]{fontenc}
\usepackage{tikz}
\usepackage{float}
\usetikzlibrary{arrows.meta, positioning, shadows}

\definecolor{myGreen}{RGB}{204, 230, 204}

\def\BibTeX{{\rm B\kern-.05em{\sc i\kern-.025em b}\kern-.08em
    T\kern-.1667em\lower.7ex\hbox{E}\kern-.125emX}}

\usepackage{hyperref}
\hypersetup{
    colorlinks=false,        
    allbordercolors={0 0 0}, 
    pdfborder={0 0 0},       
    linktoc=all              
}
\begin{document}

\title{Fed-Meta-Align: A Similarity-Aware Aggregation and Personalization Pipeline for Federated TinyML on Heterogeneous Data}

\author{
    \IEEEauthorblockN{Hemanth Macharla, Mayukha Pal\IEEEauthorrefmark{1}}\\
}

\maketitle

\begin{abstract}
Real-time fault classification in resource-constrained Internet of Things (IoT) devices is critical for industrial safety, yet training robust models in such heterogeneous environments remains a significant challenge. Standard Federated Learning (FL) often fails in the presence of non-IID data, leading to model divergence. This paper introduces \textbf{Fed-Meta-Align}, a novel four-phase framework designed to overcome these limitations through a sophisticated initialization and training pipeline. Our process begins by training a foundational model on a general public dataset to establish a competent starting point. This model then undergoes a \textbf{serial meta-initialization phase}, where it sequentially trains on a subset of IOT Device data to learn a heterogeneity-aware initialization that is already situated in a favorable region of the loss landscape. This informed model is subsequently refined in a \textbf{parallel FL phase}, which utilizes a dual-criterion aggregation mechanism that weights for IOT devices updates based on both local performance and cosine similarity alignment. Finally, an \textbf{on-device personalization} phase adapts the converged global model into a specialized expert for each IOT Device. Comprehensive experiments demonstrate that Fed-Meta-Align achieves an average test accuracy of 91.27\% across heterogeneous IOT devices, outperforming personalized FedAvg and FedProx by up to 3.87\% and 3.37\% on electrical and mechanical fault datasets, respectively. This multi-stage approach of sequenced initialization and adaptive aggregation provides a robust pathway for deploying high-performance intelligence on diverse TinyML networks.
\end{abstract}


\section{Introduction}

In the past decade, the proliferation of the Internet of Things (IoT) has embedded intelligence into the fabric of our industrial and domestic environments. This has catalyzed a paradigm shift from centralized, cloud-based data processing to on-device intelligence. At the forefront of this transformation is Tiny Machine Learning (TinyML), a rapidly advancing field at the intersection of embedded systems and machine learning. TinyML enables the execution of sophisticated ML models directly on low-cost, power-constrained microcontrollers (MCUs) with mere kilobytes of memory \cite{tinyml}. 

\begingroup\footnotesize
\thanks{(*Corresponding author: Mayukha Pal)}

\thanks{Mr. Hemanth Macharla is a Data Science Research Intern at ABB Ability Innovation Center, Hyderabad 500084, India, and also an undergraduate at the Department of Computer Science and Engineering, Indian Institute of Technology Bhubaneswar, Odisha 752050}

\thanks{Dr. Mayukha Pal is with ABB Ability Innovation Center, Hyderabad 500084, IN, working as Global R\&D Leader – Cloud \& Advanced Analytics (e-mail: mayukha.pal@in.abb.com).}
\endgroup

By processing sensor data at the source, TinyML offers profound advantages in latency, energy efficiency, bandwidth conservation, and data privacy. This is particularly transformative for applications like industrial predictive maintenance, where countless devices monitor machinery for signs of failure. The ability to detect faults in real-time on the device itself is a monumental step towards creating safer, more reliable systems. Given the sheer scale of IoT deployments, a critical research challenge is to harness the collective experience of these distributed devices. Federated Learning (FL) has emerged as a powerful framework for this task, enabling collaborative training of a global model without centralizing raw user data \cite{mcmahan2017}. In its standard form, FL protects privacy by having each device train locally and only share its model updates with a central server for aggregation. However, the theoretical elegance of FL often collides with the messy reality of real-world deployments. The data captured by each device is inherently a product of its unique environment. For fault classification, this statistical heterogeneity or non-IID (non-independently and identically distributed) data—is the norm, not the exception \cite{kairouz2019}. One motor might be prone to bearing faults, while another in a different location might experience electrical anomalies.

This heterogeneity poses a fundamental problem for traditional FL algorithms like Federated Averaging (FedAvg) \cite{mcmahan2017}, which can falter when the IOT Device data distributions diverge. A simple averaging of model updates from such diverse IOT devices can lead to a global model that is mediocre for everyone and optimal for no one, a phenomenon known as "Clientdrift" \cite{zhao2018}. This problem is further compounded in many industrial IoT settings that demand online learning, where data arrives as a continuous stream and must be processed sample-by-sample (i.e., with a batch size of one) due to severe memory constraints that prevent data storage for batch processing. Such a learning paradigm can exacerbate the instability of training on non-IID data.

To address these challenges, we introduce Fed-Meta-Align, a novel four-phase pipeline for robust federated learning and personalization, tailored for TinyML-based fault classification on heterogeneous devices. Our framework moves beyond naive averaging by first preparing a heterogeneity-aware model before collaborative training begins. It starts with a foundational pre-training stage, followed by a crucial serial meta-initialization phase to find a robust starting point. This informed model is then refined in a parallel FL phase using our similarity-aware aggregation mechanism, which intelligently weights IOT Device updates based on both local performance and cosine similarity alignment.

Recognizing that a single global model can never perfectly serve all IOT devices, the Fed-Meta-Align pipeline concludes with a lightweight personalization phase. This final step efficiently adapts the converged global model into a specialized expert for each device. This is achieved by freezing the initial half of the network's layers—preserving the robust, general features learned collaboratively—while fine-tuning the latter, decision-making layers on each device's private local data.

Our key contributions are summarized as follows:
\begin{itemize}
    \item We propose Fed-Meta-Align, a novel four-phase federated learning framework that uniquely combines serial meta-initialization with a similarity-aware parallel aggregation to effectively pre-empt and mitigate Clientdrift on heterogeneous data.
    
    \item We design a complete end-to-end pipeline tailored for TinyML, where a meta-learning phase creates a robust initialization for a more effective collaborative training phase, followed by an efficient personalization strategy that adapts the model by fine-tuning only the final layers.
    
    \item Through comprehensive experiments on a fault classification task, we demonstrate that Fed-Meta-Align substantially outperforms strong baselines, including personalized versions of both Federated Averaging (FedAvg) and Federated Proximal (FedProx).

    \item We validate the practical feasibility of our method, confirming that the final personalized models are compact, fast, and suitable for real-time, on-device inference in resource-constrained online learning scenarios.
\end{itemize}

The remainder of this paper is organized as follows. Section~II reviews related work. Section~III presents the proposed Fed-Meta-Align framework in detail. Section~IV describes the experimental setup and analyzes the results. Finally, Section~V concludes the paper and discusses future work.

\begin{figure}[t]
    \centering
    \begin{tikzpicture}[
        node distance=2cm and 1.6cm, 
        block/.style={
            rectangle, draw=black, thick, fill=myGreen, 
            text width=11em, minimum height=3.8em, text centered, 
            rounded corners, drop shadow={opacity=0.4, shadow xshift=0.1cm, shadow yshift=-0.1cm}
        },
        io/.style={
            rectangle, text width=12em, text centered, font=\small
        },
        line/.style={
            -Latex, thick
        },
        label_style/.style={
            midway, font=\small, align=center
        }
    ]

    \node[io] (data_in) {Public Fault Dataset};
    \node[block, below=of data_in] (phase0) {\textbf{Server:} \\ Foundational Pre-Training};
    \node[block, below=of phase0] (phase1) {\textbf{Server:} \\ Serial Meta-Initialization};
    \node[block, below=of phase1] (phase2) {\textbf{Server:} \\ Parallel FL Aggregation};
    \node[block, below=of phase2] (phase3) {\textbf{IOT devices:} \\ On-Device Personalization};
    \node[io, below=of phase3] (data_out) {Deployed TFLite Models};

    \node[block, right=of phase1, node distance=5.8cm, yshift=-1.4cm] (IOT devices) {IOT devices \\ \small{(Private Data)}};

    \draw[line] (data_in) -- (phase0);
    \draw[line] (phase0) -- node[label_style, right=0.2cm] {Base Model \\ $w_{base}$} (phase1);
    \draw[line] (phase1) -- node[label_style, left=0.2cm] {Meta-Initialized \\ Model $w*$} (phase2);
    \draw[line] (phase2) -- node[label_style, right=0.2cm] {Final Global \\ Model $w_{final}$} (phase3);
    \draw[line] (phase3) -- (data_out);

    \draw[line] (phase1.east) to[bend left=20] node[midway, above, font=\small, yshift=0.1cm] {Sequential Training} (IOT devices.west);

    \draw[line] (phase2.east) to[bend left=20] node[midway, above, font=\small, yshift=0.1cm] {Send Model} (IOT devices.west);
    \draw[line] (IOT devices.west) to[bend left=20] node[midway, below, font=\small, yshift=-0.1cm] {Send Updates} (phase2.east);

    \end{tikzpicture}
    \caption{The proposed four-phase architecture of Fed-Meta-Align. The framework methodically prepares the model for heterogeneity before collaboratively training and finally personalizing it for on-device deployment.}
    \label{fig:framework_overview}
\end{figure}
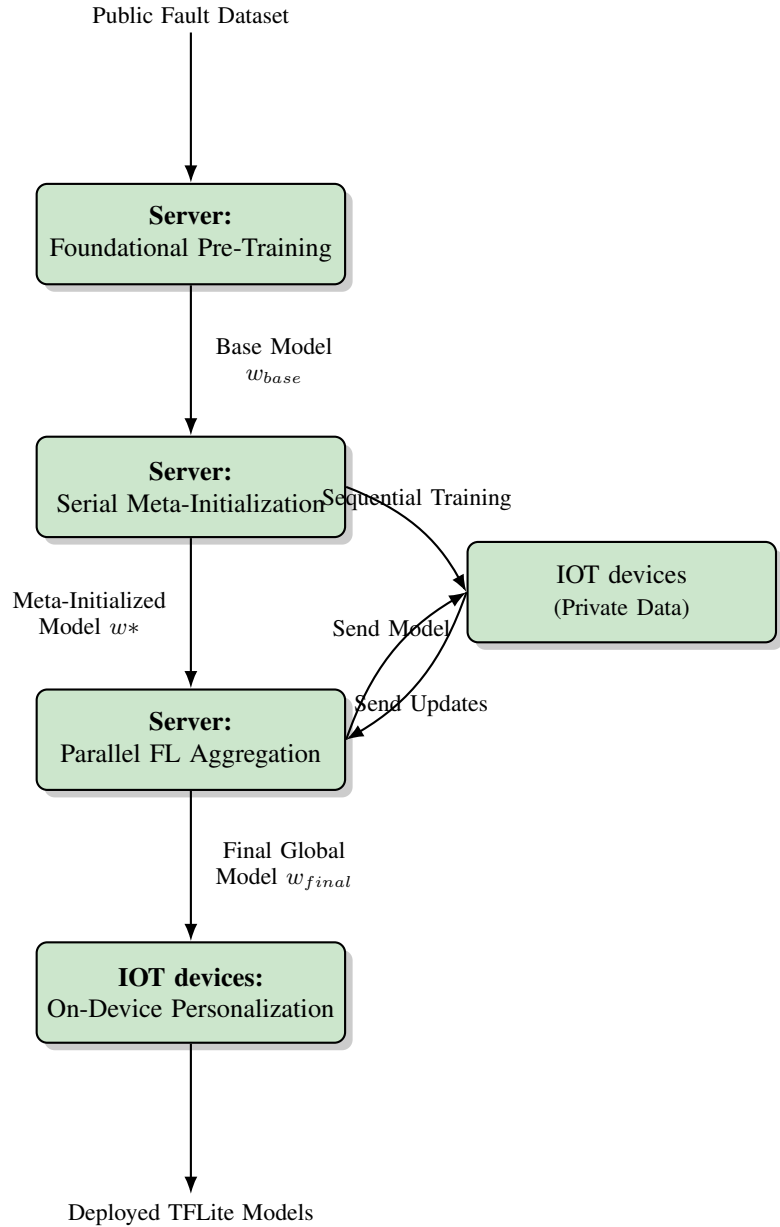

\section{Related Work}
This work introduces the Fed-Meta-Align framework, positioned at the intersection of federated learning on heterogeneous data, meta-learning for model initialization, and the practical constraints of TinyML for on-device fault classification. Key advancements in these areas are reviewed to contextualize the proposed pipeline.

\subsection{Federated Learning on Heterogeneous Data}
Federated learning enables collaborative model training across decentralized devices, with the canonical FedAvg algorithm \cite{mcmahan2017} serving as the foundational baseline. However, its effectiveness diminishes in real-world settings due to statistical heterogeneity (non-IID data), which causes local models to diverge from the global objective, a phenomenon known as client drift \cite{zhao2018}. 
To combat this, one major line of research has focused on drift correction during aggregation. For example, FedProx introduces a proximal term to the local device loss, regularizing updates to prevent divergence from the global model \cite{li2020}. Additionally, ensemble distillation techniques have been proposed to improve model robustness under non-IID conditions by aggregating diverse IoT device updates \cite{lin2020}. Personalized federated learning approaches further address heterogeneity by tailoring global models to individual device data distributions, enhancing performance on non-IID data \cite{hsu2019}. 

A parallel line of research addresses this challenge through initialization, drawing inspiration from meta-learning. The goal is to develop a global model that serves as an effective starting point, capable of rapid adaptation to each IoT device’s local data. Algorithms like Reptile demonstrate that sequential training on diverse tasks can enhance model performance \cite{nichol2018}. However, these methods are often purely serial and lack advanced aggregation strategies.

The proposed Fed-Meta-Align framework uniquely synthesizes and extends these approaches into a comprehensive, multi-stage pipeline. 
It begins with training a foundational model on a general, public dataset to establish a robust starting point with fundamental fault-related features. This is followed by a serial, meta-learning-inspired phase to create a heterogeneity-aware initialization. Finally, a similarity-aware aggregation mechanism is employed during a collaborative parallel training phase to effectively manage client drift. This sequenced approach of pre-training, meta-initialization, and adaptive aggregation provides a robust strategy for addressing real-world federated learning challenges.

\subsection{TinyML and Edge Computing}
The field of TinyML focuses on deploying machine learning intelligence on low-power microcontrollers with only kilobytes of memory. This is achieved through aggressive model optimization techniques like quantization and pruning \cite{banbury2021}. Quantization techniques, such as INT8, significantly reduce model size and inference time, enabling TinyML deployment on resource-constrained IOT devices \cite{dutta2020}. 
The intersection of FL and TinyML is a critical and emerging research area. To date, much of the work has focused on overcoming system-level challenges, primarily communication bottlenecks. However, the dual challenge of handling diverse data alongside the systemic constraints of on-device learning (e.g., small memory precluding large batch sizes) is less explored. Our framework is explicitly designed for this challenging intersection, ensuring that the final personalized models are not only accurate but also highly efficient and deployable after quantization.

\subsection{Fault Classification in IoT}
Automated fault classification is critical for the safety and maintenance of modern industrial systems. Traditional centralized models require sending all sensor data to a server, creating privacy risks and potential bottlenecks. Federated learning provides a natural privacy-preserving solution \cite{yang2019}. Federated learning also supports privacy-preserving fault detection in IoT, addressing scalability and heterogeneity challenges \cite{zhang2021}. 
However, its application to fault classification specifically for the demanding TinyML environment is still in its early stages. Our work contributes directly to this emerging field by proposing a complete, multi-stage pipeline designed to produce robust and personalized fault classification models on heterogeneous, resource-constrained devices.

\subsection{Federated Averaging and Federated Proximal}
Federated Averaging (FedAvg), introduced by McMahan et al. \cite{mcmahan2017}, is a cornerstone of Federated Learning (FL). IOT devices train local models on their private datasets, computing weight updates based on a local loss function (e.g., cross-entropy for classification). These updates are sent to a central server, which aggregates them by computing a weighted average proportional to each device's dataset size:
\[
w^{t+1} = \sum_{k=1}^K \frac{|D_k|}{\sum_{j=1}^K |D_j|} w_k^t,
\]
where \( w_k^t \) are the local weights for IoT Device \( k \) at round \( t \), and \( D_k \) is its dataset. This process iterates over multiple rounds to minimize the global loss. However, FedAvg assumes data is independently and identically distributed (IID), and its performance degrades with non-IID data, resulting in client drift \cite{zhao2018}.

Federated Proximal (FedProx), proposed by Li et al. \cite{li2020}, extends FedAvg to handle non-IID data. It adds a proximal term to the local objective, regularizing updates to stay close to the global model:
\[
\min_{w_k} \left[ \mathcal{L}_k(w_k; D_k) + \frac{\mu}{2} \| w_k - w^t \|^2 \right],
\]
where \( \mathcal{L}_k \) is the local loss, \( w^t \) is the global model at round \( t \), and \( \mu \) is a regularization parameter. This constraint reduces divergence, improving convergence in heterogeneous settings. However, FedProx addresses client drift reactively during training, unlike proactive initialization strategies.

\section{The Proposed Fed-Meta-Align Framework}

This paper introduces Fed-Meta-Align, a novel multi-stage framework designed to address the dual challenges of statistical heterogeneity and resource constraints in IoT fault classification. The pipeline is designed to methodically prepare a model for the challenges of a diverse network \textit{before} the main collaborative training even begins. The process starts by training a foundational model on a general, public fault dataset to establish a competent baseline. This base model is then refined through a serial meta-initialization process, where it sequentially 'tours' each devices data to become aware of the network's heterogeneity. With this robust foundation, the framework proceeds to a parallel federated learning stage, which uses our intelligent, similarity-aware aggregation to train a highly generalized global model. The pipeline concludes with a final, lightweight personalization step on each device, transforming the global model into a specialized expert for its local environment. The entire process is crafted with the constraints of TinyML and online learning in mind, and an overview is presented in Figure~\ref{fig:framework_overview}.

\subsection{Phase 0: Foundational Model Initialization}

The journey begins with an optional but recommended initialization phase. The primary goal here is not to solve the federated learning problem but to establish a competent baseline model. We take a standard, publicly available fault dataset and train a neural network from scratch in a centralized manner.

\begin{algorithm}[t]
\caption{Foundational Model Initialization (Phase 0)}
\label{alg:phase0_init}
\SetKwInOut{KwIn}{Input}
\SetKwInOut{KwOut}{Output}
\KwIn{A public, general-purpose fault dataset \( D_{\text{public}} \); model architecture \( A \); learning rate \( \eta \); number of epochs \( E \).}
\KwOut{Initial global model weights \( \phi_0 \) for Phase 1.}

Initialize model \( \phi \) with random weights based on architecture \( A \);\\
\For{epoch = 1, 2, \ldots, \( E \)}{
    \For{each batch \( (x, y) \in D_{\text{public}} \)}{
        Update model weights via standard optimizer: \( \phi \leftarrow \phi - \eta \nabla \mathcal{L}(\phi; (x, y)) \);\\
    }
}
Set the resulting weights as the starting point for federated learning;\\
\( \phi_0 \leftarrow \phi \);\\
\Return{\( \phi_0 \)}
\end{algorithm}

This pre-training step serves two purposes:
\begin{enumerate}
    \item Architecture Validation: It allows us to validate that our chosen neural network architecture is capable of learning the fundamental features associated with various fault types.
    \item Weight Initialization: It provides a sensible set of initial weights, \( \phi_0 \), for the central server model. Starting the federated process from this informed position, rather than from a random initialization, can help accelerate convergence in the subsequent federated training phase.
\end{enumerate}

This foundational pre-training step, summarized in Algorithm~\ref{alg:phase0_init}, is performed once, offline in a centralized manner to provide a strong set of initial weights, \( \phi_0 \), for the main federated learning phase.

\subsection{Phase 1: Serial Meta-Initialization}
This phase is an innovation of our framework, designed to transform the generic base model \( w_{\text{base}} \) into a "heterogeneity-aware" model, \( w^* \), that serves as a superior starting point for all IOT devices. This process, inspired by meta-learning, involves training the model sequentially on a reserved subset of each devices local training data. Each devices local data stream \( D_t \) is initially divided into a training set \( D_{t,\text{train}} \) and a test set \( D_{t,\text{test}} \). For this stage, \( D_{t,\text{train}} \) is partitioned into a dedicated support set \( S_{t,\text{P1}} \) for training and a query set \( Q_{t,\text{P1}} \) for evaluation.

As detailed in Algorithm~\ref{alg:phase1}, the model effectively "tours" the diverse data distributions of the training sets. In each round of this phase, the order of IOT devices is randomized to prevent any single devices data distribution from having an outsized impact on the final meta-initialized model. The model trains on a devices support set \( S_{t,\text{P1}} \), and its progress can be evaluated on the query set \( Q_{t,\text{P1}} \). Crucially, the output weights from one IOT Device become the input for the next, progressively nudging the model towards a parameter space that is a good compromise for the entire federated network.

\begin{algorithm}[t]
\caption{Serial Meta-Initialization (Phase 1)}
\label{alg:phase1}
\SetKwInOut{KwIn}{Input}
\SetKwInOut{KwOut}{Output}
\KwIn{Base model \( w_{\text{base}} \); set of IOT devices \( C \); IOT Device training data for Phase 1, \( \{D_{t,\text{train}}\}_{t \in C} \); serial rounds \( R_{\text{serial}} \); serial epochs \( E_{\text{serial}} \).}
\KwOut{Meta-initialized model weights \( w^* \).}

\( w \leftarrow w_{\text{base}} \);\\
\For{round \( r = 1, \ldots, R_{\text{serial}} \)}{
    \( O \leftarrow \text{RandomShuffle}(C) \) \tcp*{Randomize Device order each round}
    \For{each IOT Device\( t \in O \)}{
        Split \( D_{t,\text{train}} \) into support set \( S_{t,\text{P1}} \) and query set \( Q_{t,\text{P1}} \);\\
        Train \( w \) on the support set \( S_{t,\text{P1}} \) for \( E_{\text{serial}} \) epochs;\\
        Evaluate performance of \( w \) on the query set \( Q_{t,\text{P1}} \) (for diagnostics);\\
        \( w \leftarrow \) updated weights from IOT Device\( t \) \tcp*{Pass weights to next device}
    }
}
\( w^* \leftarrow w \);\\
\Return{\( w^* \)}
\end{algorithm}

\subsection{Phase 2: Similarity-Aware Federated Aggregation}

This phase is the core innovation of the Fed-Meta-Align framework. It is an iterative process, performed over multiple communication rounds, designed to build a single, generalized global model that is robust to the non-IID nature of the IOT devices' data. Let \( \phi_r \) denote the global model weights at communication round \( r \). In each round, a set of IOT devices \( C_r \) participates. The process for each IOT Device\( t \in C_r \) and the server is as follows.

\subsubsection{Local Training and Evaluation on IOT devices}
The devices remaining training data, after the allocation for Phase 1, is divided into a support set \( S_{t,\text{P2}} \) for training and a query set \( Q_{t,\text{P2}} \) for evaluation. Upon receiving the current global model \( \phi_r \) from the server, each IOT Device performs a local training and evaluation sequence utilizing the data specifically designated for Phase 2.

\begin{itemize}
    \item The IOT Device first trains the received model \( \phi_r \) on its local support set \( \mathcal{S}_{t,\text{P2}} \). With the online learning context of TinyML, this training is performed sample-by-sample. For each data point \( (x, y) \in \mathcal{S}_{t,\text{P2}} \), the model's weights are updated using the Adam optimizer with a local learning rate \( \beta \):
    \[
        \phi_{t,k+1} = \text{AdamUpdate}(\phi_{t,k}, \nabla \mathcal{L}(\phi_{t,k}; (x,y)), \beta)
    \]
    where \( \phi_{t,0} = \phi_r \). This notation signifies that the weights \( \phi_{t,k} \) are updated according to the Adam optimization algorithm. After iterating through all samples in \( \mathcal{S}_{t,\text{P2}} \), the resulting locally trained model is denoted as \( \hat{\phi}_t^{(r)} \).

    \item Next, the IOT Device evaluates this updated model \( \hat{\phi}_t^r \) on its local query set \( Q_{t,\text{P2}} \) to gauge its performance. The query loss, \( \mathcal{L}_t^Q \), is calculated as the average loss over all samples in the query set:
    \[
        \mathcal{L}_t^Q = \frac{1}{|\mathcal{Q}_{t,\text{P2}}|} \sum_{(x,y) \in \mathcal{Q}_{t,\text{P2}}} \mathcal{L}(\hat{\phi}_t^r; (x,y))
    \]

\end{itemize}

\subsubsection{IOT Device Contribution Scoring}
Instead of immediately sending the updated weights \( \hat{\phi}_t^r \) back, the IOT Device first computes metrics that will help the server intelligently aggregate its contribution.
\begin{enumerate}
    \item Query Score (Magnitude): The query loss \( \mathcal{L}_t^Q \) is transformed into a \textit{query score}, \( s_t \), which represents the empirical quality of the devices update. A lower loss indicates a better update, so we define the score as:
    \[
        s_t = \frac{1}{1 + \mathcal{L}_t^Q}
    \]
    
    This score, bounded between 0 and 1, acts as a measure of confidence in the devices update.
    \item Weight Delta (Direction): The IOT Device computes its local update vector, or weight delta, \( \Delta_t^r \), which is the difference between its updated weights and the original global weights:
    \[
        \Delta_t^r = \hat{\phi}_t^r - \phi^r
    \]
\end{enumerate}
The IOT Device then transmits both its weight delta \( \Delta_t^r \) and its query score \( s_t \) to the central server.

\subsubsection{Similarity-Aware Aggregation on the Server}
The server receives the weight deltas and query scores from all participating IOT devices. It then performs the novel similarity-aware aggregation.
\begin{enumerate}
    \item First, the server computes the average update direction across all IOT devices:
    \[
    \bar{\Delta}^r = \frac{1}{|\mathcal{C}_r|} \sum_{t \in \mathcal{C}_r} \Delta_t^r
    \]
    \item For each IOT Device\( t \), the server calculates the cosine similarity between the devices individual update direction \( \Delta_t^r \) and the average direction \( \bar{\Delta}^r \). This measures how well the devices update aligns with the collective trend.
    \[
    \theta_t = \cos(\Delta_t^r, \bar{\Delta}^r) = \frac{\Delta_t^r \cdot \bar{\Delta}^r}{|\Delta_t^r| |\bar{\Delta}^r|}
    \]
    A positive cosine similarity indicates alignment, while a negative value indicates that the IOT Device is moving in a conflicting direction.
    \item The server then computes a final, unnormalized weight \( w_t \) for each IOT Device by combining its query score (magnitude) and its directional alignment (similarity). To prevent outlier IOT devices with large negative similarities from halting progress, we bound the similarity term from below by a small constant \( c \), set to 0.1 in our experiments. This value is treated as a hyperparameter, subject to tuning based on the specific dataset and network characteristics.
    \[
    w_t = s_t \times \max(c, \theta_t)
    \]
    This ensures that even IOT devices moving in a slightly different direction can contribute, but their influence is scaled down.
    \item These weights are then normalized to sum to one, producing the final aggregation weights \( \hat{w}_t \):
    \[
    \hat{w}_t = \frac{w_t}{\sum_{j \in \mathcal{C}_r} w_j}
    \]
    \item Finally, the server updates the global model using a weighted average of the IOT Device deltas, taking a step in the aggregated direction with a server-side learning rate \( \alpha \):
    \[
    \phi^{r+1} = \phi^r + \alpha \sum_{t \in \mathcal{C}_r} \hat{w}_t \Delta_t^r
    \]
\end{enumerate}
This entire process, summarized in Algorithm~\ref{alg:fedalign}, is repeated for a fixed number of communication rounds until the global model converges.

\begin{algorithm}[t]
\caption{Parallel Similarity-Aware FL (Phase 2)}
\label{alg:fedalign}
\SetKwInOut{KwIn}{Input}
\SetKwInOut{KwOut}{Output}
\KwIn{Meta-initialized model \( w* \); IOT Device data for Phase 2, \( \{\mathcal{D}_{t,P2}\}_{t \in C} \); parallel rounds \( R_{parallel} \); server learning rate \( \alpha \); similarity floor \( c \).}
\KwOut{Final global model weights \( w_{final} \).}

\( w^{(0)} \leftarrow w* \);\\
\For{round \( r = 1, \ldots, R_{parallel} \)}{
    \For{each IOT Device\( t \in C \) \textbf{in parallel}}{
        Receive global model \( w^{(r-1)} \) from server;\\
        Split \( \mathcal{D}_{t,P2} \) into support set \( \mathcal{S}_{t,\text{P2}} \) and query set \( \mathcal{Q}_{t,\text{P2}} \);\\
        Train \( w^{(r-1)} \) on \( \mathcal{S}_{t,\text{P2}} \) to obtain updated model \( \hat{w}_t^{(r)} \);\\
        Compute query score \( s_t = 1 / (1 + \text{Evaluate}(\hat{w}_t^{(r)}, \mathcal{Q}_{t,\text{P2}})) \);\\
        Compute weight delta \( \Delta_t^{(r)} = \hat{w}_t^{(r)} - w^{(r-1)} \);\\
        Send \( \{\Delta_t^{(r)}, s_t\} \) to the server;\\
    }
    \tcp{Server-side aggregation}
    Receive \( \{\Delta_t^{(r)}, s_t\} \) from all IOT devices;\\
    Compute average delta: \( \bar{\Delta}^{(r)} \leftarrow \frac{1}{|C|} \sum_{t \in C} \Delta_t^{(r)} \);\\
    \For{each IOT Device\( t \in C \)}{
        Compute cosine similarity: \( \theta_t \leftarrow \cos(\Delta_t^{(r)}, \bar{\Delta}^{(r)}) \);\\
        Compute unnormalized weight: \( w_t \leftarrow s_t \times \max(c, \theta_t) \);\\
    }
    Normalize weights: \( \hat{w}_t \leftarrow \frac{w_t}{\sum_{j \in T} w_j} \);\\
    Update global model: \( w^{(r)} \leftarrow w^{(r-1)} + \alpha \sum_{t \in T} \hat{w}_t \Delta_t^{(r)} \);\\
}
\( w_{final} \leftarrow w^{(R_{parallel})} \);\\
\Return{\( w_{final} \)}
\end{algorithm}

\subsection{Phase 3: On-Device Personalization and TinyML Deployment}
After the federated training in Phase 2, we are left with a single, powerful, generalized model, \( \phi_{\text{global}} \). While this model is robust across diverse data distributions, it is not tailored to the specific nuances of any single device’s local environment. The final phase of our pipeline addresses this by creating a lightweight, personalized, and deployment-ready model for each device. This process leverages the remaining training data after the allocations for Phases 1 and 2, while using the test data for final evaluation and prediction. Personalized federated learning enhances model performance by tailoring global models to individual client data distributions, a principle guiding our personalization phase \cite{hsu2019}.

\begin{itemize}
    \item Personalized Fine-Tuning: The final global model \( \phi_{\text{global}} \) is sent one last time to each device. On-device, we freeze the weights of the first half of the layers (feature extraction layers) and fine-tune only the last half of the layers (decision-making layers) using the leftover training data from \( D_{t,\text{train}} \) after Phase 1 and Phase 2 splits. This adapts the model's decision-making boundary by updating the personalized model \( \phi_{t,\text{pers}} \) with a learning rate \( \gamma \):
    \[
    \theta_{t}^{\text{last}} \leftarrow \theta_{t}^{\text{last}} - \gamma \nabla_{\theta^{\text{last}}} \mathcal{L}(\phi_t^{\text{pers}}; (x,y))
    \]
    \item Threshold Optimization: To prepare the model for binary fault classification, we determine an optimal classification threshold, \( \tau_t^* \). This is done by evaluating the personalized model \( \phi_t^{\text{pers}} \) on a validation subset derived from the leftover training data and selecting the threshold that maximizes the F1-score, effectively balancing precision and recall for that specific device.
    \[
    \tau_t^* = \arg\max_{\tau} \text{F1-score}(\phi_t^{\text{pers}}, \mathcal{V}_t, \tau)
    \]
    \item Quantization and Deployment: Finally, the personalized model is optimized for efficient inference. We apply post-training quantization to convert the model’s weights to 8-bit unsigned integers (uint8), significantly reducing its size and computational requirements.
    \[
    \phi_t^{\text{quant}} = \text{Quantize}(\phi_t^{\text{pers}}, \text{target\_type=uint8})
    \]
    This quantized model is then converted into the TensorFlow Lite (TFLite) format, making it highly optimized for resource-constrained devices and MCUs. Predictions are performed using the test data \( D_{t,\text{test}} \) to assess real-world performance.
\end{itemize}

This final personalization stage, summarized in Algorithm~\ref{alg:personalization}, is executed once on each device to transform the general model into a specialized, deployment-ready expert. At the conclusion of this phase, each IoT device possesses a highly specialized, private fault classification model, ready for real-time, on-device prediction without any further need for server communication.

\begin{algorithm}[t]
\caption{On-Device Personalization and TinyML Deployment (Phase 3)}
\label{alg:personalization}
\SetKwInOut{KwIn}{Input}
\SetKwInOut{KwOut}{Output}
\KwIn{Final global model \( \phi_{\text{global}} \); fine-tuning learning rate \( \gamma \); set of all IOT devices \( C \); leftover training data \( D_{t,\text{train,remaining}} \) after Phase 1 and Phase 2 splits; test data \( D_{t,\text{test}} \).}
\KwOut{For each IOT Device\( t \in C \): a personalized model \( M_{t,\text{TFLite}} \) and threshold \( \tau_t^* \).}

\For{each IOT Device\( t \in C \) \textbf{in parallel}}{
    Receive final global model \( \phi_{\text{global}} \) from the server;\\
    Initialize personalized model: \( \phi_{t,\text{pers}} \leftarrow \phi_{\text{global}} \);\\
    Split the leftover training data \( D_{t,\text{train,remaining}} \) into a fine-tuning set \( D_{t,\text{tune}} \) and a validation set \( D_{t,\text{val}} \);\\
    Freeze the first half of the layers (feature extraction) of \( \phi_{t,\text{pers}} \);\\
    Set the last half of the layers (decision-making) of \( \phi_{t,\text{pers}} \) as trainable;\\
    \For{each data sample \( (x, y) \in D_{t,\text{tune}} \)}{
        Update only the trainable layers of \( \phi_{t,\text{pers}} \) with learning rate \( \gamma \);\\
    }
    Get model output scores on the validation set \( D_{t,\text{val}} \);\\
    Calculate optimal threshold: \( \tau_t^* \leftarrow \arg\max_{\tau} \text{F1-score}(\phi_{t,\text{pers}}, D_{t,\text{val}}, \tau) \);\\
    Convert to deployable format: \( M_{t,\text{TFLite}} \leftarrow \text{ConvertToTFLite}(\phi_{t,\text{pers}}) \);\\
}
\textbf{Procedure} Predict(\( M_{t,\text{TFLite}}, \tau_t^*, x_{\text{new}} \));\\
\( x_{\text{new}} \leftarrow \) sample from \( D_{t,\text{test}} \);\\
score \(\leftarrow\) RunInference(\( M_{t,\text{TFLite}}, x_{\text{new}} \));\\
\If{score > \( \tau_t^* \)}{
    \Return "Fault";\\
}
\Else{
    \Return "Normal";\\
}
\end{algorithm}

\section{Experimental Setup}

To validate the performance of our proposed Fed-Meta-Align framework, we conducted a series of experiments designed to evaluate its accuracy, convergence, and deployment efficiency on heterogeneous fault classification tasks.

\subsection{Datasets and Heterogeneity Simulation}
We simulate a heterogeneous IoT network using two distinct, real-world datasets, representing two IOT devices with different operational domains. A third public dataset is used for the foundational model pre-training.

\begin{itemize}
    \item Foundational Pre-training (Phase 0): We use the AI4I 2020 Predictive Maintenance Dataset to train the base model, $w$. This public dataset contains sensor data from manufacturing equipment and provides a general understanding of machine failure modes.
    
    \item IOT Device 1 (Electrical Fault data): This IOT Device uses the Electrical Fault Classification dataset. The task is to detect faults based on three-phase current (Ia, Ib, Ic) and voltage (Va, Vb, Vc) readings. The data is highly specific to electrical systems.
    
    \item IOT Device 2 (Mechanical Fault data): This IOT Device uses the Machine Failure Data dataset, which contains mechanical and environmental sensor readings like temperature, pressure, vibration, and humidity. The fault characteristics are entirely different from those of IOT Device 1, creating a significant non-IID challenge.
\end{itemize}

For each IOT Device, the data is partitioned into a training set (80\%) and a final hold-out test set (20\%). A key aspect of our experimental design is the partitioning of each devices training data to serve the distinct stages of our architecture. These partitioning ratios are treated as tunable hyperparameters. The results presented in this paper are based on a specific configuration chosen to reflect the purpose of each phase:

\begin{itemize}
    \item Serial Meta-Initialization: We allocate 20\% of the training data to this phase. The goal here is not exhaustive training but rather to allow the model to gain a foundational, heterogeneity-aware understanding by "touring" the different IOT Device data distributions.
    
    \item Parallel FL: We allocate a majority, 50\%, of the training data to this main federated learning phase. This substantial allocation ensures that IOT devices have sufficient data for robust local training across multiple communication rounds, which is essential for the convergence of the global model.
    
    \item On-Device Personalization: The final 30\% of the training data is reserved for personalization. This provides a rich, device-specific dataset for fine-tuning the global model into a specialized expert, without risking catastrophic forgetting of the collaboratively learned features.
\end{itemize}

While we believe this configuration provides a balanced and effective partitioning, a comprehensive sensitivity analysis of these ratios remains a valuable direction for future research.

\subsection{Model Architecture}
The neural network architecture used for all experiments is a standard Multi-Layer Perceptron (MLP), chosen for its suitability for tabular data and efficiency on resource-constrained devices. The model consists of:
\begin{itemize}
    \item An input layer accepting 9 features.
    \item Seven hidden layers with 256, 128, 64, 32, 16, 12 and 8 neurons respectively, each using the ReLU activation function.
    \item A final output layer with a single neuron and a sigmoid activation function for binary classification.
\end{itemize}

\subsection{Baseline Methods for Comparison}
To demonstrate the superiority of our approach, we compare it against three standard baselines:
\begin{enumerate}
    \item Local Only: A model is trained for each IOT Device using only its own local data. This baseline represents a scenario with no federated collaboration.
    
    \item FedAvg: The canonical federated learning algorithm where the server performs a simple averaging of IOT Device model weights.
    
    \item FedProx: A state-of-the-art algorithm designed to handle heterogeneity by adding a proximal term to the local loss function, regularizing local updates. We set the hyperparameter $\mu=0.01$.
\end{enumerate}
To ensure a comprehensive and fair comparison, we evaluated the baseline algorithms at two distinct stages. First, we measured the performance of the final Global Model produced by FedAvg and FedProx directly on each devices test set. This represents the performance of a standard, one-size-fits-all federated system. Second, to create a more direct and equitable comparison with our personalized framework, we also evaluated their performance After Personalization. In this step, we took the final global models from the baselines and applied the identical on-device fine-tuning process used in our own architecture before evaluating them on the local test data.

\subsection{Implementation Details}
All experiments were implemented in Python using the TensorFlow and Keras libraries. The models were trained using the Adam optimizer with a learning rate of $1 \times 10^{-5}$. For the federated phases, we ran the Serial Meta-Initialization for 10 rounds (one pass through all IOT devices per round) and the Parallel FL phase for 10 communication rounds. The final personalization phase involved fine-tuning for 10 epochs. The batch size was set to 1 to simulate an online learning environment.

\section{Results and Discussion}
This section presents and analyzes the results of the experiments conducted to evaluate the Fed-Meta-Align framework. 
The models are evaluated based on three criteria: final test accuracy, performance improvement at each stage of the architecture, and deployment efficiency for TinyML devices.

\subsection{Comparative Performance Analysis}
Our primary findings are summarized in Table~\ref{tab:accuracy_results}, which compares the final test accuracy of our proposed Fed-Meta-Align framework against the established baselines. The results provide a clear validation of our architecture's effectiveness in a challenging heterogeneous environment.

\begin{table}[t]
\centering
\caption{Final Test Accuracy (\%) on Heterogeneous IOT devices}
\label{tab:accuracy_results}
\begin{tabular}{@{}lccc@{}}
\toprule
\textbf{Method} & \textbf{IOT Device 1} & \textbf{IOT Device 2} & \textbf{Average} \\
\midrule
\textbf{Local Only} & 86.40 & 80.08 & 83.24 \\
\midrule
\textbf{FedAvg} & & & \\
\quad - Global Model & 72.34 & 54.80 & 63.57 \\
\quad - After Personalization & 84.10 & 81.60 & 82.85 \\
\midrule
\textbf{FedProx ($\mu=0.01$)} & & & \\
\quad - Global Model & 78.80 & 63.50 & 71.15 \\
\quad - After Personalization & 88.50 & 86.80 & 87.65 \\
\midrule
\textbf{Ours (Fed-Meta-Align)} & & & \\
\quad - After Personalization & \textbf{92.37} & \textbf{90.17} & \textbf{91.27} \\
\bottomrule
\end{tabular}
\end{table}

The initial results from the baseline global models starkly illustrate the challenge of statistical heterogeneity. The standard FedAvg Global Model performs exceptionally poorly, achieving an average accuracy of only 63.57\%, significantly worse than even the Local Only models that do not benefit from collaboration. This demonstrates the classic failure mode of naive federated learning, where Clientdrift leads to a converged model that is suboptimal for all participants. While FedProx mitigates this issue to some extent, its global model still underperforms compared to localized training.

The introduction of on-device personalization provides a significant performance boost across the board, establishing Personalized FedProx as our strongest baseline with an average accuracy of 87.65\%. This confirms that fine-tuning is a powerful and necessary step for adapting generalized models to specific IOT Device data distributions.

Crucially, our proposed Fed-Meta-Align framework consistently and substantially outperforms all other methods. It achieves the highest accuracy on both IOT Device 1 (92.37\%) and IOT Device 2 (90.17\%), leading to a superior average performance of 91.24\%. The improvement over the strongest baseline, Personalized FedProx, is significant: a relative gain of +3.87\% on IOT Device 1 and +3.37\% on IOT Device 2. This result validates our central hypothesis: the combination of our serial meta-initialization and similarity-aware aggregation produces a more robust and generalizable global model, which in turn serves as a far more effective starting point for the final on-device personalization.

\subsection{Analysis of Architectural Stages}
\begin{figure}[t]
    \centering
    \includegraphics[width=0.95\linewidth]{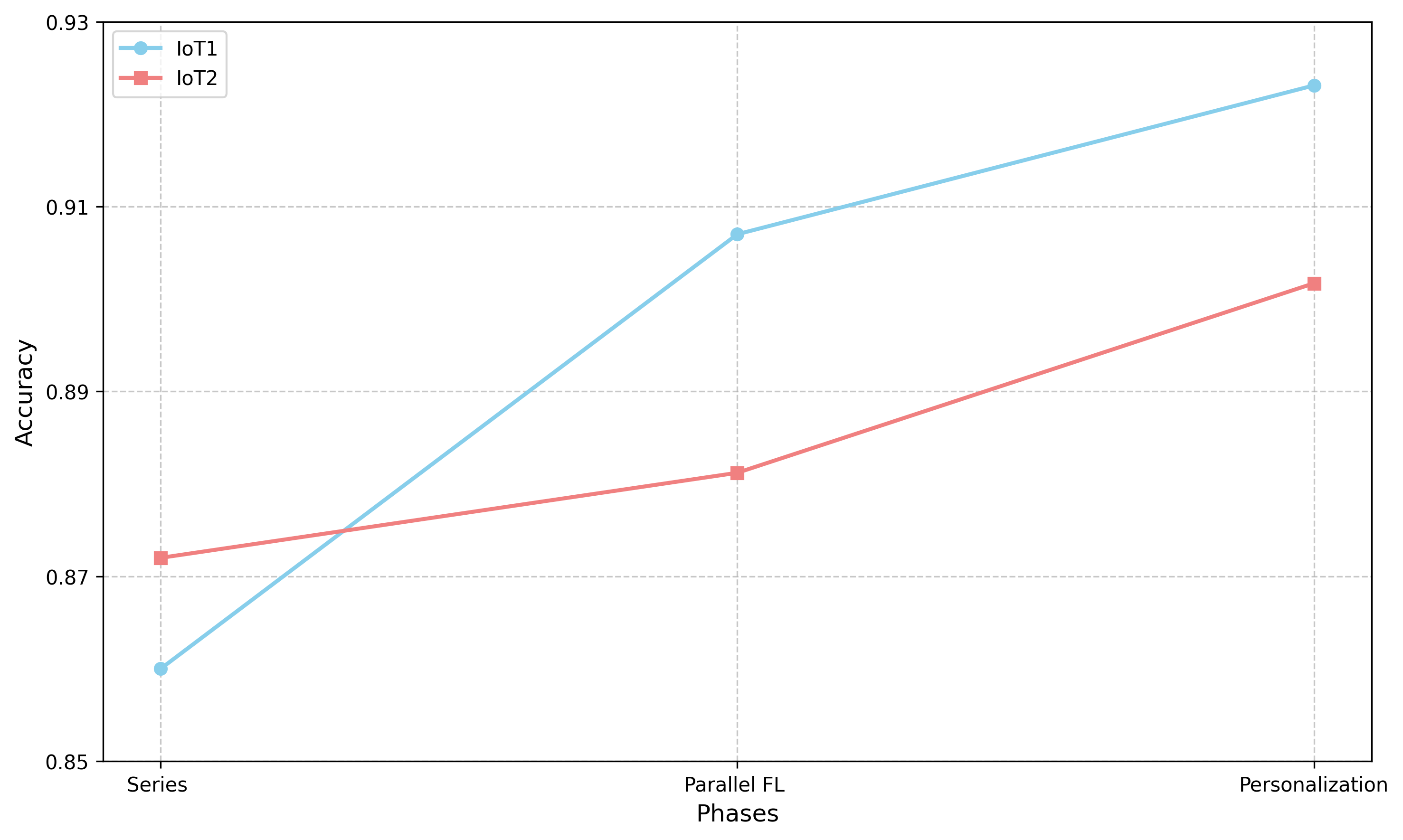} 
    
    \caption{Performance evolution for IOT Device 1 (IoT1) and IOT Device 2 (IoT2) across the three main phases of the Fed-Meta-Align framework. The graph illustrates the final test accuracy after each phase: Serial Meta-Initialization, Parallel FL, and On-Device Personalization.}
    \label{fig:stage_improvement}
\end{figure}

To validate the contribution of each distinct phase of our framework, we analyzed the performance evolution on the local test sets of both IOT devices. Figure~\ref{fig:stage_improvement} visually tracks the accuracy of the models at three key checkpoints: after the Serial Meta-Initialization phase, after the Parallel FL phase, and after the final On-Device Personalization.

The results in Figure~\ref{fig:stage_improvement} reveal a fascinating dynamic and confirm that each stage of our pipeline plays a critical and synergistic role.

Initially, after the Serial Meta-Initialization phase, we observe that the model for IOT Device 2 (IoT2) achieves a higher accuracy than IOT Device 1. This suggests that the initial "guided tour" of the data distributions results in a model that is slightly more aligned with IOT Device 2's mechanical fault data.

However, the Parallel FL phase provides the most dramatic performance lift, particularly for IOT Device 1. During this collaborative stage, IOT Device 1's accuracy shows a sharp increase, surpassing that of IOT Device 2. This is strong evidence that our similarity-aware aggregation successfully allowed IOT Device 1 to learn valuable features from IOT Device 2's updates, significantly improving its own performance. This stage effectively builds upon the strong foundation of the meta-initialized model to reach a high-quality global consensus that benefits all participants.

Finally, the On-Device Personalization stage yields the highest performance for both IOT devices. This last step acts as a crucial fine-tuning process, adapting the robust and well-generalized global model into a specialized expert for each devices specific data domain. The consistent, stepwise improvement across both heterogeneous IOT devices validates our architectural design and proves that the final high performance is not the result of a single component, but rather the collective contribution of the entire pipeline.

\begin{table}[t]
\centering
\caption{On-Device Model Footprint and Performance Comparison}
\label{tab:tinyml_results}
\begin{tabular}{@{}llccc@{}}
\toprule
\textbf{IoT Device} & \textbf{Model Format} & \textbf{Size (KB)} & \textbf{Inference Time (ms)} & \textbf{Speedup (\%)} \\
\midrule
\multirow{2}{*}{IoT Device 1} & Keras (FP32) & 183.82 & 242.00 \\
 & \textbf{TFLite (UINT8)} & \textbf{22.79} & \textbf{164.00} & \textbf{32.23} \\
\midrule
\multirow{2}{*}{IoT Device 2} & Keras (FP32) & 183.82 & 202.70 \\
 & \textbf{TFLite (UINT8)} & \textbf{22.79} & \textbf{56.30} & \textbf{72.23} \\
\bottomrule
\end{tabular}
\end{table}

\subsection{TinyML Deployment Efficiency}
A core claim of the work is the suitability of the Fed-Meta-Align framework for practical, resource-constrained TinyML applications. To validate this, we conducted a comprehensive analysis of the deployment efficiency of our final personalized models. We compared the standard Keras models (using 32-bit floating-point precision, FP32) against their 8-bit integer quantized (UINT8) TensorFlow Lite (TFLite) counterparts. The results, presented in Table~\ref{tab:tinyml_results}, measure the memory footprint (size), computational performance (inference time), and the percentage speedup achieved through quantization.

The speedup is calculated as:
\[
\text{Speedup (\%)} = \frac{T_{\text{Keras}} - T_{\text{TFLite}}}{T_{\text{Keras}}} \times 100,
\]
where \( T_{\text{Keras}} \) is the inference time of the Keras (FP32) model, and \( T_{\text{TFLite}} \) is the inference time of the TFLite (UINT8) model. This metric quantifies the reduction in inference time, reflecting the computational efficiency gained through quantization.

The data in Table~\ref{tab:tinyml_results} confirms that our approach yields models that are both compact and highly performant. The application of post-training quantization provides two critical benefits for on-device deployment.

First, the memory footprint of the personalized models is drastically reduced. For both IOT devices, the model size decreased from 183.82 KB to 22.79 KB, representing an 87.6\% reduction. This compact size makes the models exceptionally well-suited for deployment on microcontrollers with severely constrained memory resources.

Second, quantization significantly improves computational speed. For IoT Device 1, the inference time decreased from 242.00 ms to 164.00 ms, yielding a speedup of 32.23\%, as shown in Table~\ref{tab:tinyml_results}. For IoT Device 2, the improvement was even more pronounced, with inference time dropping from 202.70 ms to 56.30 ms, achieving a remarkable 72.23\% speedup. These speedups translate to approximately 1.5x and 3.6x faster inference for IoT Device 1 and IoT Device 2, respectively. The variance in speedup likely stems from differences in data complexity and the numerical properties of the learned weights for each IoT Device’s task.

\section{Inference}
This paper introduces Fed-Meta-Align, a novel four-phase framework designed for robust and personalized fault classification on heterogeneous TinyML devices. 
The approach begins by establishing a competent base model through foundational pre-training on a public fault dataset, ensuring a robust initialization. A serial meta-initialization phase creates a heterogeneity-aware starting point, followed by a similarity-aware parallel aggregation mechanism to mitigate client drift, resulting in a highly generalized global model. 
The final on-device personalization phase adapts this model into a specialized expert for each IoT device’s unique data distribution. 

The experimental results demonstrate that Fed-Meta-Align significantly outperforms standard federated learning baselines, achieving higher accuracy, particularly on IoT devices with dissimilar data. Furthermore, the final personalized models are verified to be highly efficient, with a small memory footprint and low inference latency, making them ideal for real-world TinyML deployment. This work represents a significant advancement toward building intelligent, adaptive, and scalable federated systems for next-generation IoT applications. Future work will explore the scalability of the serial phase and the application of this framework to other time-series classification tasks.


\end{document}